# Improving the Quality of Neural Machine Translation Through Proper Translation of Name Entities

Radhika Sharma[1], Pragya Katyayan[1,2], Nisheeth Joshi[1,2]
[1]Department of Computer Science, Banasthali Vidyapith, Rajasthan, India
[2]Centre for Artificial Intelligence, Banasthali Vidyapith, Rajasthan, India
sharmaradhika912@gmail.com, pragya.katyayan@outlook.com, nisheeth.joshi@rediffmail.com

*Abstract*—In this paper, we have shown a method of improving the quality of neural machine translation by translating/transliterating name entities as a preprocessing step. Through experiments we have shown the performance gain of our system. For evaluation we considered three types of name entities viz person names, location names and organization names. The system was able to correctly translate mostly all the name entities. For person names the accuracy was 99.86%, for location names the accuracy was 99.63% and for organization names the accuracy was 99.05%. Overall, the accuracy of the system was 99.52%

## I. INTRODUCTION

Machine Translation (MT) has been a hot topic of research since ages. Over time this area of study has seen paradigm shifts in the development process of MT systems. These shifts were due to the advancement in the technology space. Initially the MT system were developed using dictionaries which fetched poor results. then with the advancement of research in natural language processing (NLP), MT research gained momentum as NLP techniques (POS Tagging, Morphological Analysis, parsing etc.) were incorporated into MT pipelines. This modification drastically improved the systems, but they had their drawbacks. These were time consuming as a lot of human effort was put to develop these systems. These systems were termed as transfer/rule-based machine translation (RBMTs) systems. During 1980s, research in speech recognition was going through paradigm shift. Machine Translation was not far behind as these technologies were also being used in MT arena. This marked the rise of technologies like example-based machine translation (EBMT) system and statistical machine translation (SMT) systems.

For next few years, the MT researchers, who were working on these technologies (RBMT, EBMT, SMT) lived in congeniality until the next paradigm shift. During the start of past decade, deep learning was incorporated in machine translation research. This marked the rise of a new MT paradigm which was termed as neural machine translation (NMT).

The current state-of-the-art NMT models are performing much better than any of its predecessor technologies and each day, they are improving in performance. Tough there are still some grey areas where NMT systems are unable to provide good results. one such area is proper translation of name entities. This is because, the NMT systems have lack of understanding of context and world knowledge which leads to inconsistent translation quality. Then again, NMT systems over-relay on word-to-word translations and are extremely poor in translating text from a resource poor language. This is due to a reason that NMT models need huge amount of data to build their model upon. Since resource poor languages fail to provide such a large corpus (monolingual or parallel), they do not get good translations.

In this paper we propose a solution to solve this problem. Section 2 provides a brief literature review of the work done in machine translation. Section 3 describes our experimental setup. Section 4 elaborates on the methodology of our system. Section 5 discusses the results and section 6 concludes the work done.

## II. LITERATURE REVIEW

Liu (2015) [1] developed three types of named entity translation systems, where rule-based system didn't give satisfactory results, statistical system was average while web mining gave good results by getting correct translations of most frequent name-entities. Al-Rfou et al. (2015) [2] have demonstrated methods of building huge annotators for multiple languages with negligible human intervention. Their system builds NER annotators for 40 major languages with the help of Freebase and Wikipedia. Sulea et al. (2016) [3] investigated the utility of word embeddings for the task of named entity translation from a resourceful language to a low-resource language. They introduced a way to obtain bi-lingual word vectors. Shao (2016) [4] presented transliteration system which was based on deep learning techniques developed for normal tasks of sequence processing using CNN for character level information extraction. The system works for Chinese-English and vice versa.

Li et al. (2017) [5] studied the technique that performs named entity segmentation as sub-words or characters. They used neural network based seq2seq mechanism to build a transliterator. They applied this to transliterate personal names over LDC dataset. Jain and Agarwal (2017) [6] explained two methods of developing transliteration tool that converts English text to Sanskrit. First approach is by using a physical keyboard for the purpose while second approach involves the use of a virtual keyboard. Grundkiewicz and Heafield (2018) [7] built a strong system for transliteration by applying successful NMT techniques like model ensemble, back translation, drop-out regularization and right-to-left models.

Merhav and Ash (2018) [8] analyzed the challenges affecting the development of transliteration system for named entities in multiple languages. They evaluated encoder-decoder RNN methods as well as non-sequential transformer techniques. Ameur et al. (2018) [9] built an open-source named entity transliteration model and identification dataset for English-Arabic. Yan et al. (2019) [10] studied the possibility of applying external named entity translation model to NMT. They experimented by implementing several named entity translation models with NMT models and reported the impact of this structure in enhancement of translation results. Moran and Lignos (2020) [11] evaluated LSTM, GRU, bi-LSTM and transformer architectures for named-entity transliteration in a multi-lingual setup. Their system was capable of working from 590 languages to English using various combinations of encoder-decoder combinations and evaluated them on the criterion of character rate, accuracy and F-score based on longest continuous subsequences.

Mon et al. (2020) [12] constructed a named entity dictionary for Myanmar-English having 80k transliteration objects. They used statistical and neural methods where neural outperformed the statistical model on the basis of Bleu score. Li et al. (2021) [13] proposed a mixed method of named entity translation for bi-lingual entity class based on the strategy of chunk symmetry. They also built an English-Chinese transliteration model using machine learning. Mon and Soe (2021) [14] proposed named entity terminology lexicon for Myanmar-English. They used transformer-based neural models. They used BLUE score to evaluate neural model's performance. Mota et al. (2022) [15] proposed an approach to handle named entities for improving quality of translation in an NLP pipeline. They implemented separate steps of recognizing named entities and translation.

### III. PROPOSED SYSTEM

Since we were translating the name entities, we needed a resource for identifying name entities. Thus, we used Stanford's name entity recognizer (NER) for this purpose. Further, we also used an English-Hindi parallel corpus from which we had manually extracted the name entities. Next, we wrote an algorithm for extracting phonemes from both English as well as Hindi name entities and applied word-to-world alignment on them using Berkley Aligner. Further, we had also developed a knowledge base where most of the organization names and location are translated from English to Hindi.

The phoneme extraction was done through a set of rules where the characters in phonemes had two classes (vowels and consonants). Any phoneme would have either a vowel or a consonant with a vowel or some complex combination of the two classes. Some of the examples of these are shown in table 1.

The algorithm for extraction of the same is as follows:

**Algorithm 1: Phonify**

**Input:** word

**Output:** phonemes

**Function Phonify:**

{a,e,i,o,u} are stored as list of vowels and remaining characters are stored as list of consonants.

Consider a vowel and its next consonant as a phoneme

If a consonant is followed by a vowel than consider them as phoneme

If consonant is followed by more than one vowel than consider them as separate phoneme

Consider all other combinations as separate phoneme.

TABLE I. LIST OF PHONEME STRUCTURES WITH EXAMPLES

| Phoneme Structure | Example Phoneme | English (Phonified) | Hindi (Phonified) |
|---|---|---|---|
| V | Amar | **[A]** [ma] [r] | **[अ]** [म] [र] |
| CV | Radhika | **[Ra]** [dhi] [ka] | **[रा]** [धि] [का] |
| VC | Anshika | **[An]** [shi] [ka] | **[अं]** [शी] [का] |
| CCV | Odisha | [O] [di] **[sha]** | [ओ] [डी] **[शा]** |
| CVC | Cherapunji | [Che] [ra] **[pun]** [ji] | [चे] [रा] **[पुं]** [जी] |

Once the English and Hindi phonemes are extracted, their probability table is generated for implementation of hidden Markov model (HMM) as it is a very handy mechanism in identifying context and disambiguating accordingly. Here two types of probabilities are calculated. First is the transition probability which calculates probability of a Hindi phoneme based on previous/another Hindi phoneme. Second is the likelihood probability where we calculate the probability of English phoneme based on Hindi phoneme. These transition and likelihood probabilities are calculated using equations 1 and 2 respectively. Further based on these two types of probabilities three segments are calculated. These are transition probability of current phoneme based on previous phoneme, transition probability of next phoneme based on current phoneme and likelihood probability of current English phoneme based on current Hindi phoneme. These three form the basis for HMM algorithm and is shown in equation 3. Table 2 shows a snapshot of phoneme probability calculation.

$$P(E_i|H_i) = \frac{Freq\ (H_iE_i)}{Freq\ (H_i)} \quad (1)$$

$$P(H_i|H_{i-1}) = \frac{Freq\ (H_{i-1}H_i)}{Freq\ (H_{i-1})} \quad (2)$$

$$P(H_i|E_i) = P(E_i|H_i) \times P(H_i|H_{i-1}) \times P(H_{i+1}|H_i) \quad (3)$$

For translating English text into Hindi, the first step was to identify if the text was a name entity. if the extracted name entity is an organization name or location, then search for their exact names in knowledge base. If found, then generate the output else extract phonemes from the name entities. This was done because most of the times in compound name entities we find organization names (in some cases location names) as different in English and Hindi. Table 3 gives examples of such cases where we have difference in English and Hindi organization names. Transliterating them through phonification algorithm makes the text less fluent. For example, let us consider a sentence

India is a great country.

इंडिया एक महान देश है। *(disfluent)*

भारत एक महान देश है। *(fluent)*

It has just one name entity which is India. Transliterating it through phonification in Hindi will fetch इंडिया while the correct Hindi translation should be भारत. Thus, a sentence with भारत is more fluent.

Using the Hidden Markov Model, we generated equivalent Hindi phoneme. Finally, merged the phonemes and generated the output text in Hindi. Figure 1 shows the entire working of the system.

TABLE II. SNAPSHOT OF PHONEME PORBABILITY TABLE

| English Phoneme | Hindi Phoneme | Probability |
|---|---|---|
| ra | रा | 0.894 |
| m | म | 0.5491 |
| ba | ब | 0.6921 |
| t | त | 0.5987 |

TABLE III. ORGANIZATION NAMES

| Name Entities (English) | Name Entities (Hindi) |
|---|---|
| Indian Institute of Technology | भारतीय प्रोद्योगिकी संस्थान |
| Finance Ministry | वित्त मंत्रलाया |
| Indian Railways | भारतीय रेल |
| Central Secretariate | केन्द्रीय सचिवालय |

## IV. EVALUATION

For evaluating the system, we collected 1000 sentences which had 29283 name entities in them. We restricted our focus on only three types of name entities viz person, location, and organization. Among the 29283; 11713 were person names, 7320 were location names, and 10250 were organization names.

The system was able to correctly transliterate 11697 person names which gave an accuracy of 99.86%. for location names, it was able to generate 7293 correct translations, thus giving an accuracy of 99.63%. For organization names, the system was able to correctly translate 10153 entities, giving an accuracy of 99.05%. overall, 29143 name entities were correctly translated/transliterated. Thus, giving an overall accuracy of 99.52%. Table 4 gives the summary of this study.

TABLE IV. SUMMARY OF EVALUATION

| Name Entities | Total | Correct | Accuracy |
|---|---|---|---|
| Person | 11713 | 11697 | 0.99863 |
| Location | 7320 | 7293 | 0.99631 |
| Organization | 10250 | 10153 | 0.99053 |
| All Three | 29283 | 29143 | 0.99521 |

## V. CONCLUSION

In this paper, we have shown the development of a name entity translation module which translates or transliterates name entities. For translation we have created a knowledge base which performs rule-based matching. For transliteration, we have developed a phonification algorithm which extract phonemes from the name entities and then applies HMM algorithm to correctly identify the Hindi equivalent of the English name entity. This process also helps in disambiguating phonemes which have more than one match. The system was evaluated for person names, location names and organization names. The accuracy achieved for all the name entities was around 100%.

As a future extension of this study, we would like to add more name entities and extend our knowledge base and probability table so that the system can further improve its performance.

ACKNOWLEDGMENT

This work is supported by the funding received from SERB, GoI through grant number CRG/2020/004246 for project entitled, "Development of English to Bharti Braille Machine Assisted Translation System".

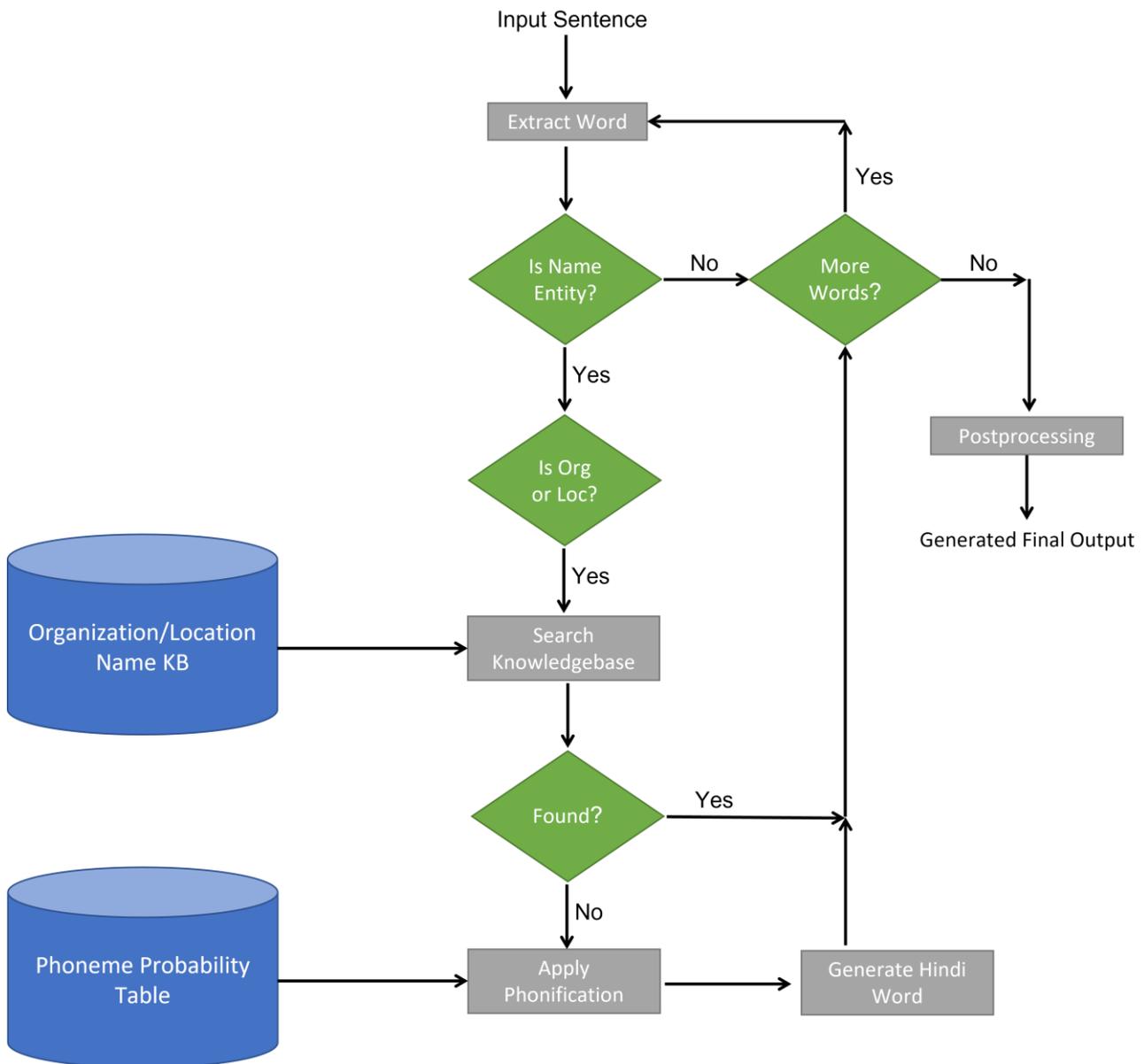

FIGURE 1    WORKING OF NAME ENTITY TRANSLATION SYSTEM.